\documentclass[10pt,twocolumn,letterpaper]{article}

\usepackage{cvpr}
\usepackage{times}
\usepackage{epsfig}
\usepackage{graphicx}
\usepackage{amsmath}
\usepackage{amssymb}
\usepackage{makecell}

\usepackage[breaklinks=true,bookmarks=false]{hyperref}

\cvprfinalcopy 

\def\cvprPaperID{****} 

\begin{document}

\title{UTS submission to Google YouTube-8M Challenge 2017}


\author{Linchao Zhu\hspace{4em}Yanbin Liu\hspace{4em}Yi Yang\\
	University of Technology Sydney\\
	{\tt\small \{zhulinchao7, csyanbin, yee.i.yang\}@gmail.com } 
}

\maketitle


\begin{abstract}
In this paper, we present our solution to Google YouTube-8M Video Classification Challenge 2017.
We leveraged both video-level and frame-level features in the submission.
For video-level classification, we simply used a 200-mixture Mixture of Experts (MoE) layer, which achieves GAP 0.802 on the validation set with a single model.
For frame-level classification, we utilized several variants of recurrent neural networks, sequence aggregation with attention mechanism and 1D convolutional models.
We achieved GAP 0.8408 on the private testing set with the ensemble model.
  The source code of our models can be found in \url{https://github.com/ffmpbgrnn/yt8m}.
\end{abstract}


\section{Introduction}
The basic methodology towards untrimmed video classification can be 1) frame-level/clip-level feature generation; 2) leveraging video context information; 3) temporal aggregation.
In the YouTube-8M dataset, two frame-level features are provided, which are static image features extracted by the Inception network~\cite{szegedy2016rethinking} pre-trained on ImageNet
and audio features extracted by a VGG-inspired acoustic model~\cite{hershey2016cnn} trained on the first version of YouTube-8M.
The original testing videos were not available during the competition, and new features could not be extracted.
We thus focus on 2) and 3) in the paper.
First, each frame is conditioned on the previous frames and the orders of the frames need to be leveraged.
We then utilize aggregation methods which eliminate order information but abstract discriminative representations for classification.
We first present our approach in Section~\ref{sec:approach} and the results are shown in Section~\ref{sec:expr}.
The conclusion is drawn in Section~\ref{sec:conclusion}.


\section{Our Approach}
\label{sec:approach}
We first show our initial analysis of the dataset. We then present our approach using video-level and frame-level features.

\subsection{Dataset Analysis}
Videos have \textbf{multiple labels}. To calculate the loss, a simple method is to replace the softmax function with the sigmoid function.
Second, we can use the softmax function but the labels need to be smoothed.
Third, regarding the video tag assignment problem in a sequence to sequence scenario,
label prediction can be generated at each decoding step with a softmax function. There is no order relation between the labels but we manually sort the labels in the vocabulary order.
From our preliminary results, the sigmoid function always performs best.

\begin{figure*}[t]
  \centering
  \includegraphics[width=0.98\linewidth]{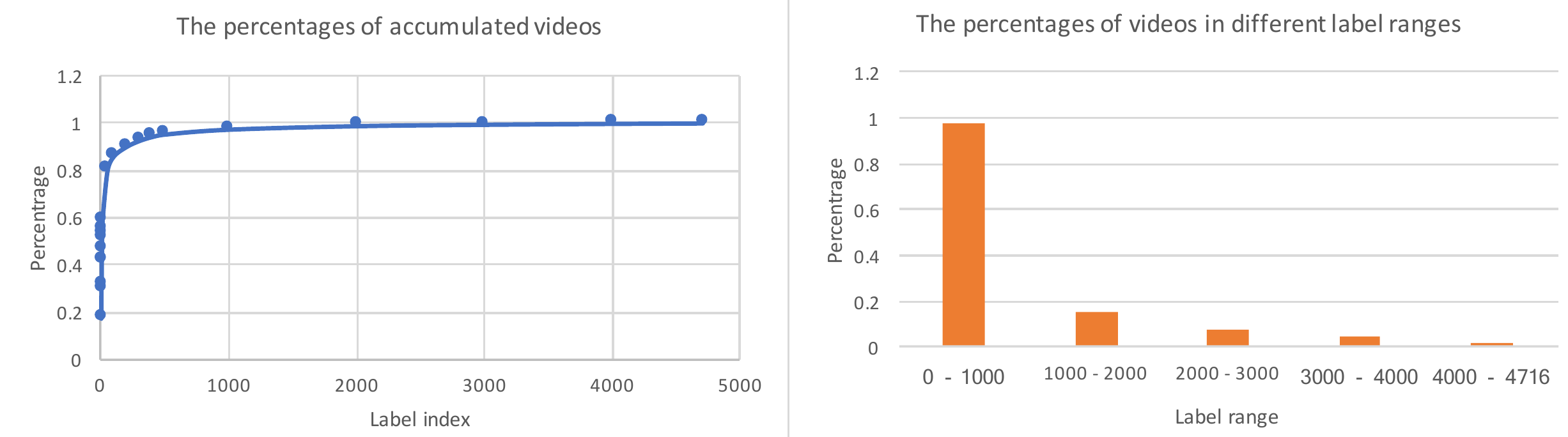}
  \caption{The vocabulary is basically ordered with regards to the number of positive instances in the class, \ie, smaller label indices have more positive instances.
  We can observe that most videos (97\%) are in label range [0, 1000). In label range [0, 50), there are 80\% of the videos, while less than 1\% videos are in [3000, 4716).
  }
  \label{fig:label_distribution}
\end{figure*}

\begin{figure}[t]
  \centering
  \includegraphics[width=1.0\linewidth]{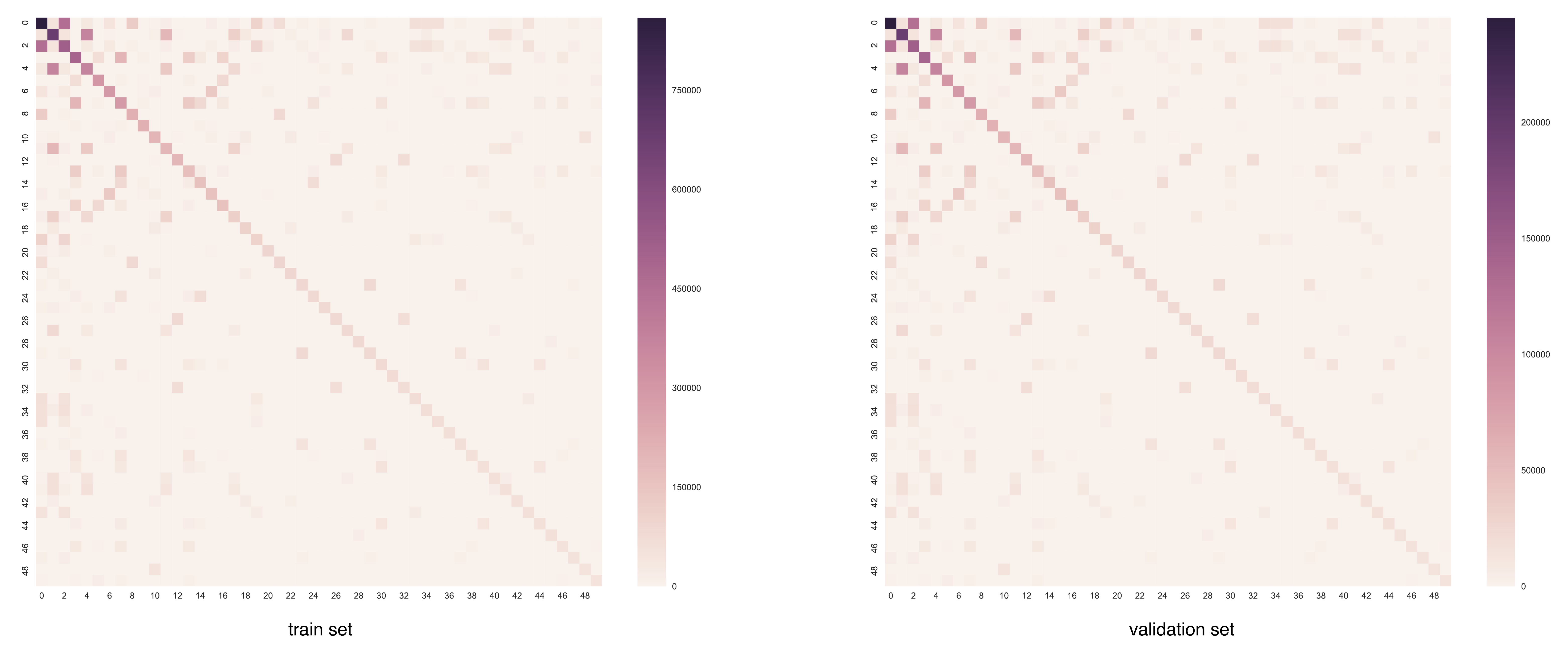}
  \caption{Label co-occurrence matrix of top 50 labels on the training set (left) and the validation set (right).}
  \label{fig:co_occur}
\end{figure}

The YouTube-8M dataset is \textbf{imbalance} that some categories (excluding top-level categories) have over 50k positive examples,
\eg, ``Minecraft'', while some categories have only 100 positive examples, \eg, ``Mammal''~(Figure~\ref{fig:label_distribution}).
We tried to keep the label balanced in a minibatch during training,~\ie, the number of positive examples are similar for the categories in a minibatch. However, it results in worse performance.
We also tried to normalize the label weight with its frequency,~\ie, higher frequency labels have lower weights, but it does not help either.
Our explanation is that the videos are imbalanced in the training, validation and testing set, and the loss calculated by random sampling estimates the ``true'' distribution better.
In Figure~\ref{fig:co_occur}, we show the label co-occurrence matrix on the training and validation set, which have very similar distribution.

The YouTube-8M dataset is also \textbf{noisy}. Abu-El-Haija~\etal~\cite{abu2016youtube} reported that the precision and recall of labels are about 78.8\% and 14.5\%, respectively. Missing and noisy data are common in this dataset. To tackle this issue, we attempted to remove some noisy labels or complete the missing labels for each class in a certain amount, \eg, 5\%. Both methods have no influence on the performance.
We plan to investigate this in the future.
We did not deal with label noise or use positive negative sampling in the following models.


\subsection{Video-level Classification}
In video-level classification, one feature vector is provided for each video.
\subsubsection{Mixture of Experts (MoE)}
Given input $x$, the mixture of experts layer is directly applied on the input, which is calculated by,
\begin{equation}\begin{aligned}
  f_{moe}(x) = \sum_{i=1}^{k}{G(x)_iE_i(x)},
\end{aligned}\end{equation}
where $G(x)_i$ is the gating weight for expert $i$, and $E_i(x)$ is the prediction of the $i$th expert.
We adopted the MoE layer used in~\cite{abu2016youtube}, where sigmoid activation is used on the expert output and the gating weights are soft assigned with a softmax function.

\subsubsection{Parallel MoE (PMoE)}
In our preliminary experiments, we found that increasing the number of mixtures from 2, 4 to 8 will increase the classification performance.
We aim to train MoE with hundreds of mixtures.
\cite{shazeer2017outrageously} used a sparsely-gated MoE with thousands of mixtures where only a small number of mixtures are updated in the training.
Another way is to use hierarchical MoE~\cite{shazeer2017outrageously}.
In the multi-label classification setting, we can simply use model parallelism that the vocabulary is divided into small label groups,
where one MoE layer only predicts a subset of the whole vocabulary.
There are different ways to divide the vocabulary. One way is to divide the labels in vocabulary order.
For example, in the 200-mixture setting, we can divide the labels into ranges \{[0, 500), [500, 1000),~\ldots, [4500, 4716)\}.
We can also randomly sample non-overlap labels from the vocabulary.

\subsection{Frame-level Classification}
In frame-level classification, the inputs are in variable lengths, which are zero-padded to sequence length 300.
We use RNNs to model temporal transitions, while attention or VLAD are for aggregation.
1D ResNet is also used.

\begin{figure*}[t]
  \centering
  \includegraphics[width=0.8\linewidth]{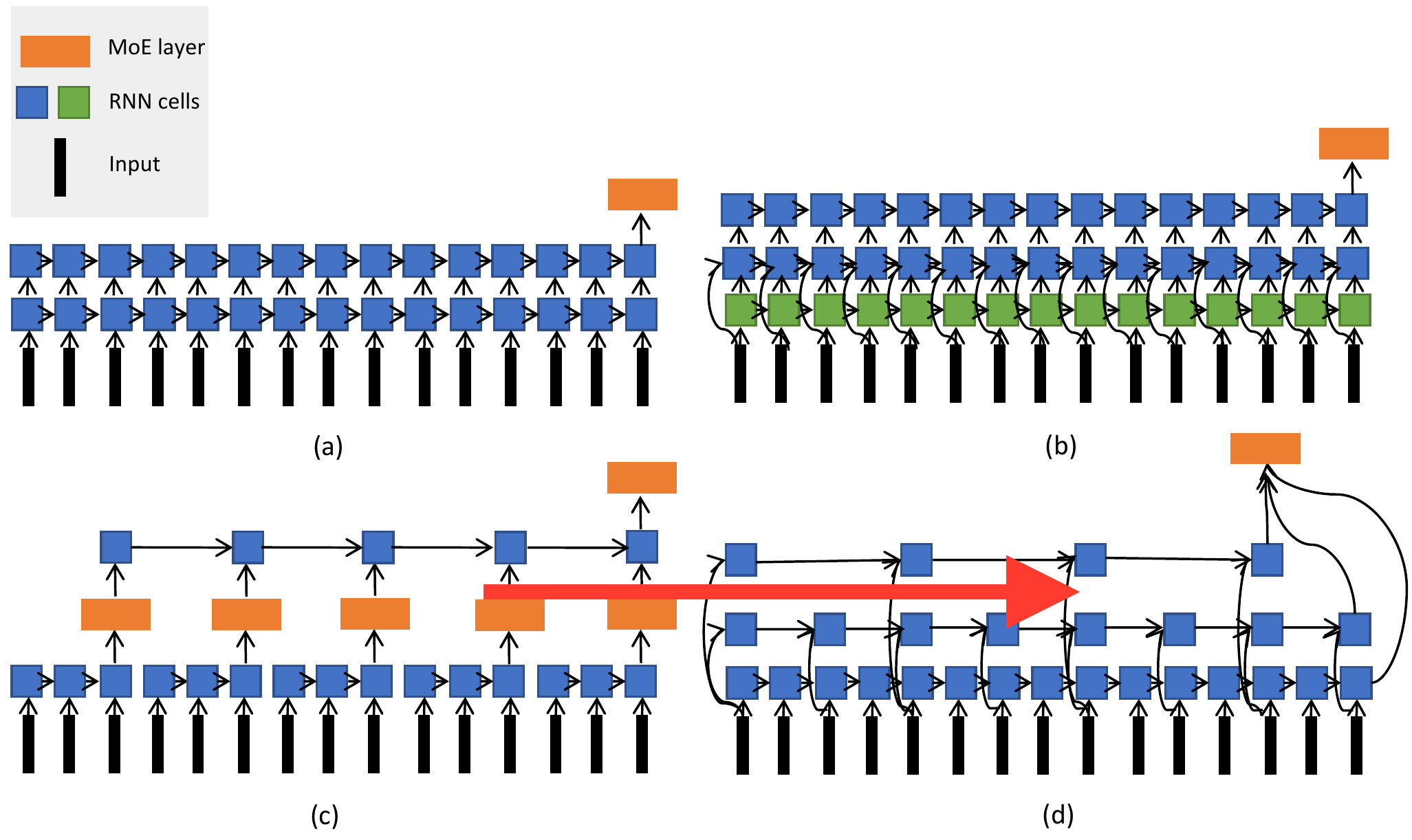}
  \caption{RNNs variants used in our submission. (a) Stacked RNN; (b) Stacked RNN with context injection; (c) HRNN; (d) Multi-scale RNN.}
  \label{fig:rnn_variants}
\end{figure*}

\subsubsection{Recurrent Neural Networks}
RNNs have been successfully applied to video classification~\cite{ng2015beyond,icml2015_srivastava}, where Long Short-Term Memory (LSTM)~\cite{lstm} and
Gated Recurrent Unit (GRU)~\cite{Cho_GRU} are commonly used.
We use the following variants in our submission (Figure~\ref{fig:rnn_variants}):

a) \textbf{Two-layer stacked RNN}. We stack two layer RNNs and evaluate the performance of different RNN architectures,~\eg, GRU and LSTM, on this setting.
A MoE layer is added on the RNN output for classification.

b) \textbf{Stacked RNN with context injection}. Following~\cite{zhu2016bidirectional}, we use a seq2seq model to reconstruct video contexts, where an encoder encodes a clip to reconstruct its previous and next clips.
The encoder thus encodes information beyond the seen clip. We design the stacked RNN model which takes $x$ and the outputs of the context encoder as inputs, which is

\begin{equation}\begin{aligned}
  x = \text{stop\_gradient}(\text{ContextRNN}(x)) + x.
\end{aligned}\end{equation}
Note that we do not backpropagate the gradient through the context encoder.

c) \textbf{Hierarchical RNN with hidden MoE layer}. One problem with the stacked RNN is that it is computation expensive and the gradient may still vanish when the sequence length grows.
Following~\cite{pan2015hierarchical}, we use a hierarchical RNN where the first RNN encodes video segments information and the second RNN further aggregates each segment.
We plug a MoE layer between the layers but no activation funcion is used on the expert outputs.

d) \textbf{Multi-scale RNN}. In this variants, we divide the inputs into several groups with different intervals. The original frames are sampled at 1FPS, we further subsample the frames with lower frame rate.
States of different rates are then concatenated for classification.



\subsubsection{\textbf{Vector of Locally Aggregated Descriptors}}
\begin{figure}[t]
  \centering
  \includegraphics[width=0.98\linewidth]{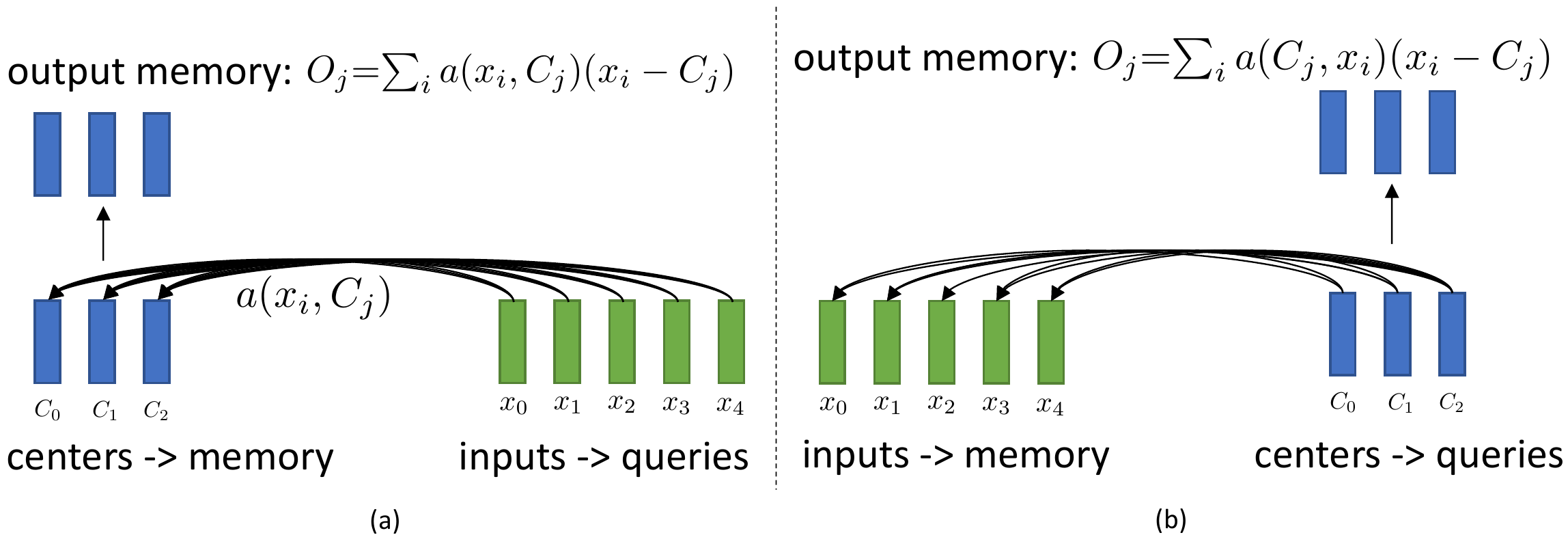}
  \caption{Two schemes to calculate the kernel $a$. The difference is that in (a), softmax is calculated over centers while in (b) it is calculated over the inputs.}
  \label{fig:vlad_attns}
\end{figure}

Instead of using the final state/output for classification, we can also use a weighted sum over the outputs of all steps.
Linear annealing weights~\cite{ng2015beyond} and attention with multiple hops can be used.
We modified the attention mechanism as follows,

\begin{equation}\begin{aligned}
  output = flatten(softmax(\mathbf{W}_atanh(\mathbf{W_i}\mathbf{X}^T))\mathbf{X}),
\end{aligned}\end{equation}
where $\mathbf{X}$ is the input with shape [$\text{seq\_length}$, $\text{input\_size}$],
$\mathbf{W_i}$ is the projection matrix with shape [\text{proj\_size}, \text{input\_size}],
$\mathbf{W_a}$ is the attention matrix with shape [\text{num\_hops}, \text{proj\_size}].

Vector of Locally Aggregated Descriptors (VLAD) can aggregate frame-level ConvNets activations for video classification~\cite{xu2015discriminative}.
The length of the feature vector is usually over 60,000 which is also very sparse. Thus training classifiers,~\eg, logistic regression, on VLAD can easily lead to overfitting.
Training VLAD end-to-end has been first attempted in~\cite{arandjelovic2016netvlad} and later Girdhar~\etal~\cite{girdhar2017actionvlad} applied NetVLAD to action recognition.
The original VLAD is calculated by, 
\begin{equation}\begin{aligned}
\label{eq:vlad}
  O(j) = \sum_{i=1}^{N}{a(x_i, C_j)(x_i - C_j)},
\end{aligned}\end{equation}
where $x_i$ is the $i$th input, $C_j$ is the $j$th center, $a(x_i, C_j)$ denotes
the membership of the input $x_i$ to center $C_j$, \ie, $a(x_i, C_j)=1$ if center $C_j$ is closest to input $x_i$ and 0 otherwise.
Instead of hard assignment,~\cite{arandjelovic2016netvlad,girdhar2017actionvlad} used a soft assigment with
\begin{equation}\begin{aligned}
  a(x_i, C_j) = \frac{\exp\{-\alpha||x_i-C_j||^2\}}{\sum_{k'}{\exp\{-\alpha||x_i-C_{k'}||^2\}}},
\end{aligned}\end{equation}
which is further decoupled into,
\begin{equation}\begin{aligned}
  a'(x_i, C_j) = \frac{\exp\{w_j^Tx_i+b_j\}}{\sum_{k'}{\exp\{w_{k'}^Tx_i+b_{k'}\}}}.
\end{aligned}\end{equation}
We use another kernel $a(x_i, C_j)$ that is commonly used in attention,
\begin{equation}\begin{aligned}
  a''(x_i, C_j) = \frac{\exp\{W_atanh(W_cC_j+W_ix_i+b)\}}{\sum_{k'}{\exp\{W_atanh(W_cC_{k'}+W_ix_i+b)\}}}.
\end{aligned}\end{equation}
Another constraint is added to minimize $\sum_{i=1}^{N}{a(x_i, C_j)||x_i - C_j||^2}$.

Equation~\ref{eq:vlad} is very similar to the attention machnism with two differences. First, Equation~\ref{eq:vlad} has an additional residual connection. Second,
the weighted sum is applied on the memory in attention, while in the VLAD case, the weighted residuals are concatenated. We would investage more in the future.
We show two different schemes to calculate the kernel $a$ in Figure~\ref{fig:vlad_attns}.

\subsubsection{1D Convolution}
By replacing the 2D kernel with 1D kernel and keeping other setups unchanged, we train a 1D ResNet-50~\cite{he2016deep} on the features provided.
The input ``1D image'' has shape [300, 1].
As the length of the input channel is 1,536, we increase the first convolutional channel to 512, and the following channels are 512, 1,024, 2,048, 4,096.
Global average pooling is applied and softmax activation is replaced by a sigmoid activation.

\subsection{Fusion}
\label{subsec:Fusion}
In our submission, for each class, we fuse the predictions weighted by the Average Precision (AP) score on the validation set.
We normalize the per-class APs of all models as the class-level fusion weights.
Given $M$ predictions, and the APs on the validation set for $i$th prediction are ${\text{ap}_{i,0}, \text{ap}_{i,1}, \dots, \text{ap}_{i,4715}}$, %
we can calculate the weights for class $c$ over all models by
\begin{equation}
    W_c = \text{norm}(\text{ap}_{0,c}, \text{ap}_{1,c}, \dots, \text{ap}_{M-1,c}).
\end{equation}
The normalization function can be $\ell_1$-norm, $\ell_2$-norm or other norms. 


\section{Experiments}
\label{sec:expr}

Two features are provided for each frame sampled at 1FPS. We did not investigate late fusion and the two features were concatenated as inputs to all the models.
We trained the models with the following settings unless otherwise stated.
We optimized the models with ADAM optimizer and the learning rate decays 0.9 every 4,000,000 examples.
The initial learning rate for MoE models is 0.01, and we use learning rate 5e-4 when training RNNs.
GAP is the evaluation metric and we report both the GAP and the mAP scores on the validation set.
The metrics are reported using single checkpoint for each model.


\subsection{Video-level Classification}
We show the results of the MoE models in Table~\ref{table:video_cls_moe}. The performance increases when the number of mixtures increases.
For 2, 4, 8, 50, 100 mixtures, we used a single MoE layer and we trained on CPU if OOM occurred on GPU.
For the 200-mixture model, we divided the vocabulary in order with window size 500, while the labels are randomly selected in ``200random''.
We can see the performance difference between ``200'' and ``200random'' is small.
For the 1000-mixture model, we only trained on the first 1,000 labels. We averaged the prediction of
200-mixture and 1000-mixture which achieves GAP 0.8141 on the validation set.

\subsection{Frame-level classification}
We first show the results of our stacked RNN variants on Table~\ref{table:frame_cls_stackrnn}.
By default, we used 2 stacked layers and 2-mixture MoE for classification.
From the result, we can see that LSTM performs worse than GRU and bidirectional RNN performs slightly worse than the basic RNN.
Besides, increasing the number of mixtures did not boost the performance. In ``GRU+fc'', we added an output projection on the GRU states.

In Table~\ref{table:frame_cls_otherrnn}, we show the result of other RNNs.
``HGRU'' is the hierarchical GRU with the default window size 15.
We used 2-mixture MoE and dropout keep ratio is set to 0.5. ``LN HGRU'' is the HGRU
where the activations are layer normalized~\cite{ba2016layer}, which does not have improvements.
We observed over-fitting in these models and thus shuffled the order of input frames (``HGRU (random order)''), but it leads to worse performance.
``Multi-scale GRU'' has similar performance to ``HGRU''.
A slightly improvement can be obtained by changing the windows size from 15 to 20 (``HGRU@20'').

The results of 1D ResNet and our modified end-to-end VLAD are shown in Table~\ref{table:frame_cls_others}.
Note that in these two models, the inputs are the original frame-level features rather than outputs of RNNs.
The initial learning rate is set to 0.1 for 1D ResNet and it decays 0.1 every 10 million examples.
In our preliminary experiments, we can obtain accuracy 82.28\% (rgb only) on UCF-101 split 1 using our version of VLAD, where the average pooling result is 78.57\%.
Notably, only 10 center is used in UCF-101 and the dimension of the feature vector is 2,560.
We used 10 centers on YouTube-8M and did not try other settings. Further investigation would be made in the future.

\begin{table}
\begin{center}\begin{tabular}{|c|c|c|}
\hline
    \# of Mixtures &  mAP       &  GAP      \\
\hline\hline
    2                 &  0.4150    &  0.7820   \\
\hline
    4                 &  0.4180    &  0.7890   \\
\hline
    8                 &  0.4205    &  0.7932   \\
\hline
    50                &  0.4262    &  0.8000   \\
\hline
    100               &  0.4249    &  0.8011   \\
\hline
    200               &  0.4291    &  0.8023   \\
\hline
    200random         &  0.4291    &  0.8019   \\
\hline
    200+1000          &  \textbf{0.4430}    &  \textbf{0.8141}   \\
\hline
\end{tabular}\end{center}
\caption{The results of different number of mixtures in MoE for video-level classification.}
\label{table:video_cls_moe}
\end{table}

\begin{table}
\begin{center}\begin{tabular}{|c|c|c|}
\hline
    Models             &  mAP     &   GAP    \\
\hline\hline
    LSTM              &  0.3884  &  0.8044  \\
\hline
    BiLSTM            &  0.4035  &  0.8042  \\
\hline
    GRU               &  \textbf{0.4302}  &  \textbf{0.8147}  \\
\hline
    BiGRU             &  0.4280  &  0.8092  \\
\hline
    GRU+4-mixture     &  0.4258  &  0.8127  \\
\hline
    GRU+fc            &  0.4187  &  0.8128  \\
\hline
\end{tabular}\end{center}
\caption{The results of stacked RNN variants. In this case, we found increasing the number of mixtures in MoE did not improve the performance.}
\label{table:frame_cls_stackrnn}
\end{table}

\begin{table}
\begin{center}\begin{tabular}{|c|c|c|}
\hline
    Model             &  mAP       &   GAP      \\
\hline\hline
    HGRU              &  0.4429    &   0.8243     \\
\hline
    LN HGRU           &  0.4385    &   0.8220  \\
\hline
    HGRU (random order) &  0.4078  &   0.8150  \\
\hline
    Multi-scale GRU        &  \textbf{0.4464}   &   0.8225  \\
\hline
    HGRU@20            &  0.4445   &  \textbf{0.8246}  \\
\hline
    StackGRU+context (*)           &  -   &  0.8210  \\
\hline
\end{tabular}\end{center}
\caption{The results of other RNN variants. * denotes the model is not fused in the final submission.}
\label{table:frame_cls_otherrnn}
\end{table}

\begin{table}
\begin{center}\begin{tabular}{|c|c|c|}
\hline
    Model                       &  mAP      &   GAP      \\
\hline\hline
    1D ResNet                   &  0.4294   &  0.8176 \\
\hline
    AttnVLAD+2-mixture          &  0.4282   &  0.8004 \\
\hline
\end{tabular}\end{center}
\caption{The results of 1D ResNet and AttnVLAD. The ``AttnVLAD'' model is our modified end-to-end VLAD.}
\label{table:frame_cls_others}
\end{table}

\subsection{Fusion}
In our final submission, we fused all the above models to obtain the prediction. Some models were not used in the final submission, and we did not report their results.
The results on validation and test (private) can be seen in Table~\ref{table:frame_cls_hrnn}. $\ell_3$-norm fusion obtained the highest results on validation set and this submission got GAP 0.84081 on the test set (private).

\begin{table}
\begin{center}\begin{tabular}{|c|c|c|}
\hline
    Model             &     GAP (val)      & GAP (private) \\
\hline\hline
    Average Fusion    &    0.840819   &  - \\
\hline
    $\ell_1$ APs            &    0.841035   & - \\
\hline
    $\ell_2$ APs            &    0.841142   &  -\\
\hline
    $\ell_3$ APs            &    \textbf{0.841169}   & \textbf{0.84081} \\
\hline
\end{tabular}\end{center}
\caption{The fusion results. By simply averaging all prediction, we can obtain GAP 0.840819. Slight improvement can be achieved with our fusion method.}
\label{table:frame_cls_hrnn}
\end{table}


\section{Conclusion}
\label{sec:conclusion}

In this paper, we presented our solution to YouTube-8M Challenge. We found hierarchical GRU model with MoE has the best single model performance, and Multi-scale GRU performs slightly worse.
Motion information is missed but we would like to evaluate how much the motion information contributes to the performance on the untrimmed noisy data in the future.
Another future work is to evaluate the aggregation methods on the large-scale dataset through an ablation study.


{\small
\bibliographystyle{ieee}
\bibliography{yt8m}
}

\end{document}